# Flying By ML - CNN Inversion of Affine Transforms

L. Van Warren
*Computer Science*
*University of Arkansas*
Little Rock, USA
lvwarren@ualr.edu

*Abstract*— This paper describes a machine learning method to automate reading of cockpit gauges, using a CNN to invert affine transformations and deduce aircraft states from instrument images. Validated with synthetic images of a turn-and-bank indicator, this research introduces methods such as generating datasets from a single image, the 'Clean Training Principle' for optimal noise-free training, and CNN interpolation for continuous value predictions from categorical data. It also offers insights into hyperparameter optimization and ML system software engineering.

## I. Introduction

Cockpit instrumentation is essential for monitoring aircraft state and making control decisions. While modern "glass cockpits" use digital displays, many traditional analog gauges remain due to their reliability and cost. This work explores using machine learning (ML) and computer vision techniques to deduce aircraft states by processing images of cockpit instruments.

Specifically, a convolutional neural network (CNN) approach is developed to invert affine image transformations like rotation, scaling, and translation to extract instrument readings. This enables interfacing robotic systems with existing cockpit gauges without costly upgrades or modifications that might compromise safety.

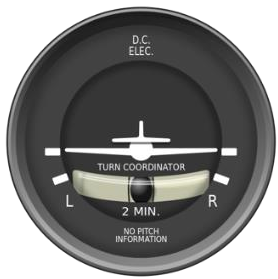

Fig. 1. A Standard Turn-and-bank Indicator

A turn-and-bank indicator displaying aircraft roll angle is used for validation. From a single image, train/test sets are synthesized by affine transformations. The CNN deduces roll angles accurately, achieving 100% test accuracy and recognizing angles from -90 to +90 degrees in 1-degree increments. These limits and increments can be reduced, extended, or refined without significant programming effort.

## II. Background

The historical development of autopilots, from Sperry's early mechanical systems to modern automated glass cockpits, reflects the trajectory of increasing automation and pilot support systems, suggesting a potential role for ML in enhancing aviation safety. This shift towards advanced systems begins in the modern era with autopilot technology in Piper and Mooney M20 aircraft and continues in general aviation [1]-[6].

Furthermore, this work draws on historical and operational contexts, such as the human factors considered in transitioning to glass cockpits, the function of the turn coordinator instrument, and the characteristics of light aircraft like the Cessna 172. Practical insights into roll dynamics are gained from aerobatic demonstrations and foundational research on roll control [7]-[11].

This CNN-based approach promises to fulfill aviation's stringent demands for real-time performance, safety, and robustness under diverse flight conditions, which have been limitations in prior meter reading research.

Combining the evolution of autopilot technology with advancements in machine learning, this dissertation centers on applying CNNs to read analog cockpit instruments, a critical step towards enhancing automation in general aviation. While digital "glass cockpits" are becoming the norm, the retention of analog instruments for redundancy underscores the importance of developing cost-effective ML techniques to interpret these gauges [12]-[14].

In summary, the work integrates historical insights, operational knowledge, and technical advancements, underpinning the development of a CNN-based wing leveling system for light aircraft as its canonical example, mirroring the evolution begun by Sperry.

SEVEN INNOVATIONS

## A. Hybridization of Old and New Technologies Rather Than Compulsive Obsolescence

The wing leveling portion of the autopilot problem has been solved for nearly a century but has yet to be adequately solved with the aid of a CNN. Combining long-solved problems with novel deep-learning neural network technologies is fruitful because one can compare and contrast the development and deployment of solutions over time. Solutions that solve old problems in new ways illuminate their similarities and differences. In this way, new hybrid approaches emerge that release previous solutions and approaches from restrictions imposed by historical technology limitations.

## B. Inversion of Affine Transformations

Affine Transformations are commonly implemented as 4x4 matrices, which, when multiplied by a spatial coordinate, can rotate, scale, translate, skew, and apply perspective to it. This applies to geometric points in 4D spacetime and pixels in video image streams. Affine coordinate transformation is the bread and butter of modern 3D computer graphics hardware and software [18]- [21].

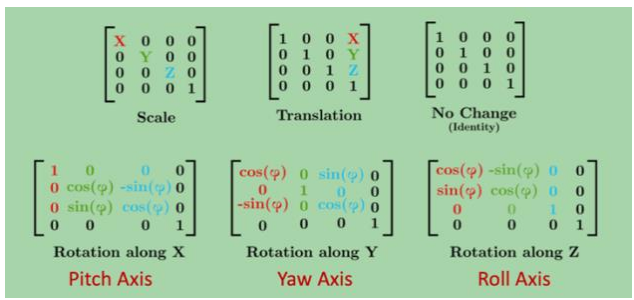

Fig. 2. 4x4 Affine Transforms in Roll, Pitch, and Yaw

It is most commonly the case that the transformations are applied in the forward direction where a control point that is part of a larger geometry is scaled, rotated, translated, skewed, or placed in perspective. In this work, we are going the other way, where given a transformed object, we ask for the parametric value that gave rise to it. In this case, that parameter is roll angle.

## C. Biomimetic Sensing Reduces Complexity and Expands Functionality (VFR to IFR)

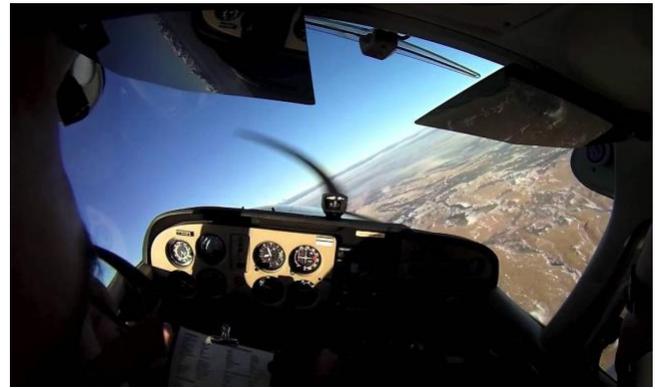

Fig. 3. Releasing the Assumption of VFR (fair weather) Conditions

When using ML as an autopilot for roll control, placing the video sensor outside the aircraft was the first design impulse. The plan was to train the CNN to recognize all kinds of flight attitudes, terrain circumstances, lighting conditions, and weather. This would have required collecting extensive training footage that was general enough to represent combinations of all these circumstances.

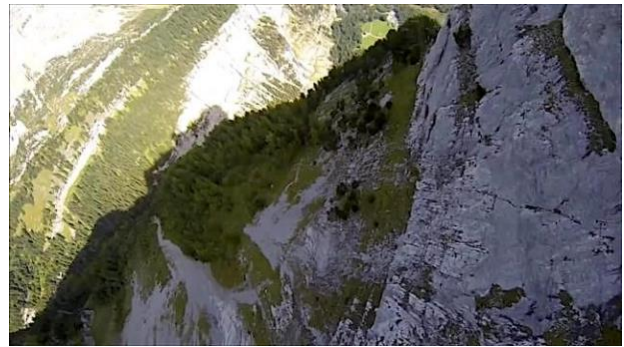

Fig. 4. We're flying. Where is the horizon?

For example, imagine that one is flying by the side of a mountain, and the video sensor thinks the incline of the mountainside represents level ground. This is a problem. The solution to which illuminates an important robotic principle. Instead of pointing the camera at the external scene in all its glaring generality, we point it at an instrument, a specialized sensor that depicts our orientation in space. Though simple to state in retrospect, the emergent principle is most important. By directly imaging the dial face of a gyroscopically driven instrument, a windfall occurs, enabling the solution of four crucial problems:

1. A vast reduction in visual and computational complexity. No need to digitize all flight scenarios.
2. A solution that applied to all-weather flight, VFR, and IFR.
3. There is no need to tamper with or replace expensive, existing, proven instrumentation.
4. Specialized sensors reduce ML workload and H/W requirements by orders of magnitude.

As stated earlier, this work started assuming that the external sky and horizon were visible from the aircraft's cockpit (A/C).

In that case, the CNN must be trained for all conceivable exterior scenes and weather conditions, which is a massively complex problem. A second limitation of using external scenery was that this limited the scope of this work to flight under FAA Visual Flight Rules (VFR). The focus was sharpened, and the assumptions were released by realizing that if the Turn-and-Bank Indicator was visible to a cockpit camera, then sensing could be done under FAA Instrument Flight Rules (IFR), which allow flight in more varied and severe weather circumstances. The IFR technique, the superset of flight regulations, wins because the instrument continues functioning without external visual cues or videographic information from outside the cockpit. This creates a much more robust sensing and control circumstance and dramatically reduces the number of states the ML program must recognize. The key observation was that the readout from an instrument has a finite number of states that can be reduced via image processing to a small catalog of instrument states, which lends itself immediately to an MNIST-like approach for the near-instantaneous determination of aircraft state of roll. In effect, rather than trying to solve the general-purpose robot vision problem including foggy weather, we are equipping our robot with a one-axis vestibular system, an 'inner ear' that provides a sense of balance for rotation about the roll axis. By using a CNN, we can also immunize the sensing to noise.

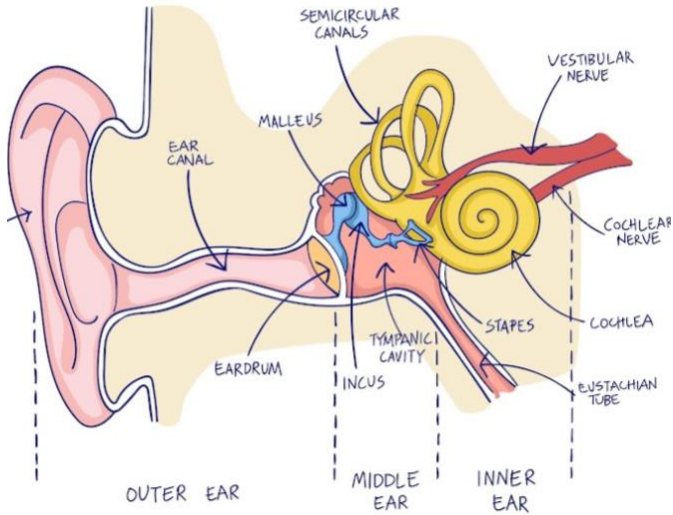

Fig. 5. The Human Vestibular System has Sensors for Roll, Pitch, and Yaw

Amplifying this point further, our robotic sensing imitates the biological human vestibular system shown in the figure above. We could sense with our vision, but even nature noticed that this creates a significant and distracting cognitive overload. Instead, we determine our geometric attitude with a specialized sensor perfectly analogous to the human (and animal) semicircular canals, one for each roll, pitch, and yaw axis. We were made to fly!

### D. Test/Train Datasets Generated from Single Keyframe

The train and test sets can be generated by the forward affine transform of a single keyframe into a set of discrete states the instrument is expected to register. This saves labor in data collection and cleaning.

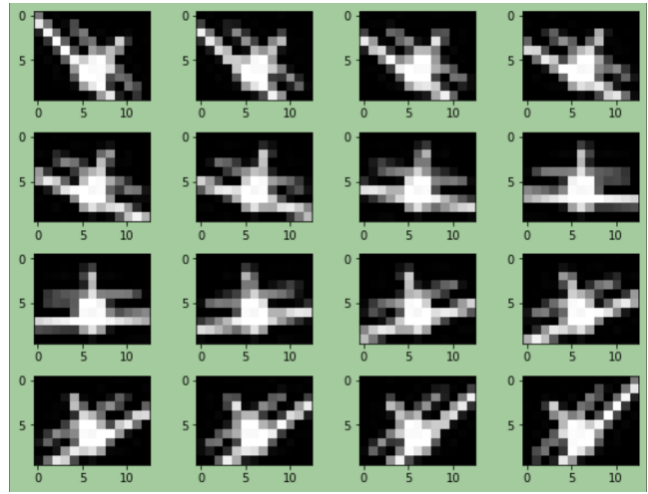

Fig. 6. Test/Train Set from Single Keyframe

### E. The Clean Training Principle and Functioning in the Presence of Noise

This work also considered the impact of noise on the training of the CNN.

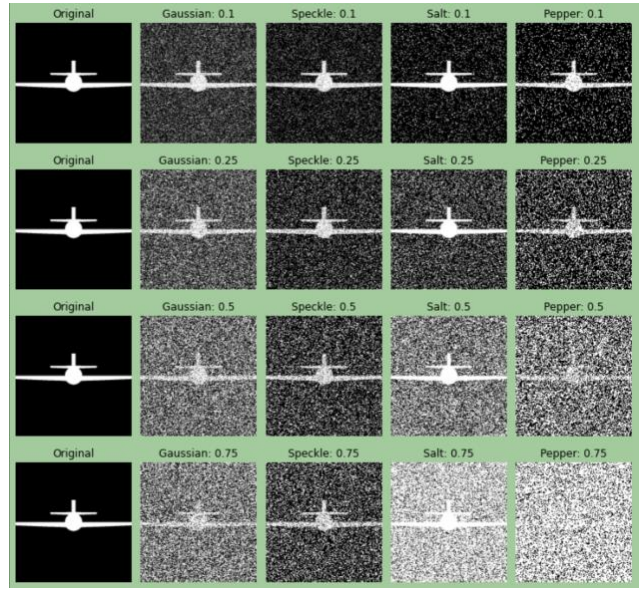

Fig. 7. A Variety of Noise Levels & Types Can Be Imposed During Training

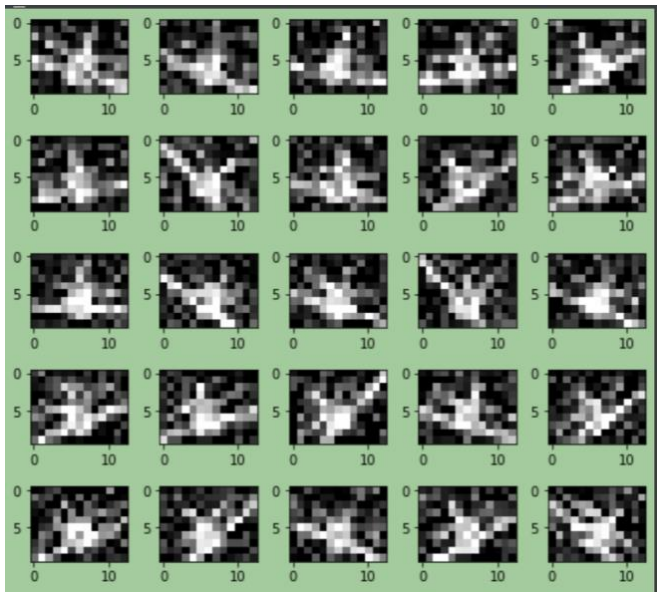

Fig. 8. Training and Test Data with Added Noise

"The Clean Training Principle" was articulated during the process, showing that training without noise is superior to training with noise. Initially, noise training was believed to prevent overfitting the CNN model.

### F. Artificial Interpolation with a CNN

A well-designed CNN can improve the accuracy of instrument reading approximations by interpolating values between those provided in the training set. This work extends classical CNNs to enable interpolation using a dot product with the entire decision vector, allowing all elements to participate in the final decision. This is as opposed to the traditional method of using the *most likely candidate* computed via the softmax() or argmax() functions. This "artificial interpolation" approach can be more accurate than the number of samples would seem to allow. This is covered in detail in the section below.

### G. Hyperparameter Optimization

This is discussed in the Technical Approach section below and more thoroughly in its section.

In summary, this work demonstrates a practical approach for robotic systems to safely interface with existing cockpit instrumentation using computer vision and deep learning. This also applies to dials, instruments, and gauges on the factory floor, which may have similar legacy constraints.

### III. CNN-BASED TECHNICAL APPROACH

First, a short review of the literature:

Yalçın's history of CNNs offers a narrative arc from biological roots to contemporary models, clearly explaining key concepts but omitting technical specifics. LeCun et al.'s groundbreaking paper presents the LeNet-5 architecture, a cornerstone for CNNs in document recognition, with notable strengths in architecture and training. The MNIST dataset, detailed by LeCun & Cortes, remains a touchstone for CNN research, primarily focusing on simple digit recognition [22] - [24].

Gupta provides actionable insights for achieving high accuracy in MNIST classification, with optimization strategies applicable to a broader context. Brownlee's comprehensive guide on machine learning optimization encompasses critical methods like hyperparameter tuning, informing the performance enhancements in your model. Patel's visually engaging video makes CNN concepts accessible to novices, although it does not delve into the deeper mathematics of the field [25] - [27].

Kalinin's thorough explanation of CNN mathematics demystifies the mechanics of CNNs despite covering a partial breadth of backpropagation. Lastly, Wang et al.'s interactive tool demystifies CNN operations through hands-on exploration, despite its static nature [28] - [29].

This work develops a CNN-based approach using the TensorFlow ML library to train and test the CNN using synthesized aircraft instrument data. The work is written in Python and runs in the cloud using Google Drive for file management, accessing the plentiful GPU resources on Google Collab, and using Jupyter notebooks with vi editor extensions as the development platform [30] – [34].

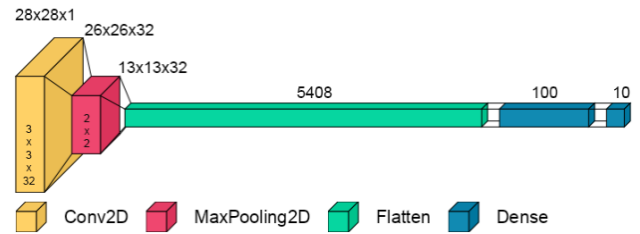

Fig. 9. MNIST Model Layers

The workflow consists of:

A. Image capture

B. Dataset creation

C. CNN training

D. Hyperparameter optimization

E. Framewise inferencing and prediction

### A. Image Capture

A simulated Cessna 172 cockpit in the X-Plane 11 flight simulator provides photo-realistic instrument imagery. A single 1036 x 1036 RGB image captures the composite turn-and-bank indicator. Image processing extracts the key moving element, reducing it to a 68,200-pixel grayscale glyph. This approach could be applied without modification by furnishing an actual or synthesized image of the roll indicator for any aircraft, including electronic ones. Splitting the imagery into static parts (the dial face) and dynamic parts (the moving indicator) is a simple, one-time image-processing task.

*B. Dataset Creation*

Train and test datasets are synthesized by affine transforming this glyph over +/- *max_bank_angle* in n-degree increments. For a typical training case, mage mirroring produces left/right bank angles for a 20-image "canon". This work examined duplicating this test canon as many as 6000 times, yielding 120,000 synthesized training examples. These angle ranges and increments are parameterized to obtain arbitrary accuracy. Coarse increments significantly speed the CNN train, test, verification, and hyperparameter optimization steps. After this, the coarse increments are replaced with finer ones, enabling better than 1-degree accuracy using artificial interpolation - an invention of this work described previously.

*C. CNN Training*

A CNN model inspired by MNIST handwriting recognition is developed with two significant differences. The images of subsequent rotation of the indicator glyph form an ordered set, making artificial interpolation possible. The standard MNIST digits are ordered but possess no geometric regularity defining that order, binding them forever to the categorical interpretation. This work's CNN has two convolutional layers extracting 128 feature maps, then dense layers mapping features to a *num_bank_angles* layers output predicting bank angle categories. With a mature hyperparameter configuration, only five epochs of training on the synthesized dataset are required to achieve 100% training accuracy. On all fronts, using the MNIST precedent helps to ensure a method that satisfies aviation safety requirements, which are stringent.

*D. Hyperparameter Optimization*

Systematic experiments reveal optimal hyperparameters that balance accuracy and efficiency. Adding training examples beyond 6000 copies provides negligible accuracy benefit. Five epochs are sufficient for convergence. Image sizes can be reduced up to 22X with no accuracy loss. This enables faster training and inference. It is worth noting that CNNs have incredible visual acuity, allowing a significant reduction in glyph size while maintaining acceptable accuracy [35].

*E. Framewise Inferencing and Prediction*

The trained model deduces aircraft bank angle from individual instrument images in real-time at over 1000 frames per second, more than sufficient for real-time control decisions. Artificial interpolation significantly increases effective resolution beyond the fixed number of training examples and should be explored further. It is an example of type conversion from category to number and imitates human visual estimation during measurements.

IV. ARTIFICIAL INTERPOLATION WITH A CNN

*A. An Encouraging Discovery*

It was initially believed that we must represent every possible dial face we would ever want to recognize in the training set. This was not the case and is a significant result of this work. For an analog instrument such as the bank indicator, an infinite number of states must be discretized into a reasonable number of digital states that enable aircraft control. An encouraging discovery made during this process was that the CNN can interpolate undersampled bank angles when they are dotted with the prediction vector weights returned by the CNN, as opposed to the typical softmax()/argmax() function called in MNIST to select the recognized digit from the prediction vector. The only requirement for this to work is that the glyph processed by the CNN must not be symmetric across both the horizontal and vertical axis, which is an easy requirement to satisfy, as developed above. Said another way, the glyph can never be such that it aliases to other shapes when transformed under rotation, translation, or scaling. This limits angle values to the principal values of +/- 180-degree banks, which are perfectly adequate for this purpose.

Examples generated for the training set must be such that there are no 'missing codes' in the range of angles the dial will be read. 'Missing codes' is a term drawn from the performance analysis of Analog to Digital converters (ADCs). An ideal ADC can reproduce evenly-spaced digital codes corresponding to the entire range of analog inputs. The CNN designed here is, in effect, an ADC spanning an air gap between the instrument and the converter. This has other security and surveillance possibilities that will not be explored here.

The next step is to do artificial interpolation with a CNN to invert the affine transformation that indicates the aircraft's roll angle. This is this work's second mathematically interesting result and possibly the one with the broadest applicability. As explained above, it is called artificial because we interpolate categories of related objects to produce a numerical rather than a categorical result. This is detailed as follows:

*B. Artificial Interpolation via Prediction Vector Trick*

The CNN is adapted to recover the roll angle using the entire prediction vector. This inverts the order of the usual process of affine transformation. This procedure is analogous to the 'kernel trick' in Support Vector Machines**Error! Reference source not found.**.

*C. Categorical vs Numerical Results*

We usually think of CNNs as giving categorical results instead of numerical ones, e.g., Car, Truck, Bicycle vs. 30°, 60°, and 90° by the category with the most significant value probability.

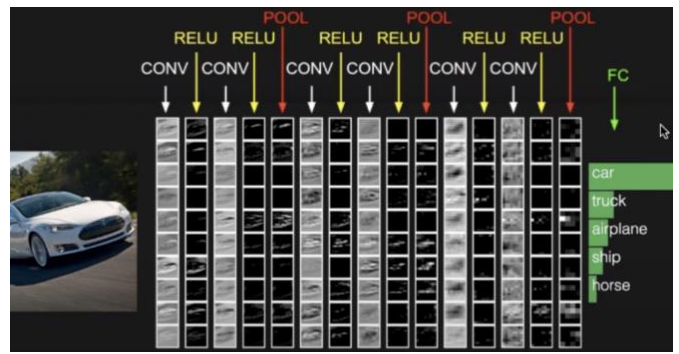

Fig. 10. CNNs Typically Give Categorical Results

*D. Numerical vs. Category Result Enabled Using Dot Product*

```
# compute categorical bank angle
categorical_bank_angle_idx = np.argmax(prediction_vector)
# compute categorical bank angle
categorical_bank_angle_idx = np.argmax(prediction_vector)
```

Fig. 11. Numerical vs. Category Result via Dot Product

Here, we align the vector of possible bank angles with the results of the prediction vector in preparation for the dot product step:

```
bank_angle_array:
  -90  -87  -84  -81  -78  -75  -72  -69  -66  -63
  -60  -57  -54  -51  -48  -45  -42  -39  -36  -33
  -30  -27  -24  -21  -18  -15  -12   -9   -6   -3
    0    3    6    9   12   15   18   21   24   27
   30   33   36   39   42   45   48   51   54   57
   60   63   66   69   72   75   78   81   84   87
   90
prediction_vector:
    0    0    0    0    0    0    0    0    0    0
    0    0    0    0    0    0    0    0    0    0
    0    0    0    0    0 0.08 0.92 0.00    0    0
    0    0    0    0    0    0    0    0    0    0
    0    0    0    0    0    0    0    0    0    0
    0    0    0    0    0    0    0    0    0    0
    0
actual    image    angle    is: -13
predicted argmax symbol is: #26
predicted argmax angle   is: -12
artificial interp angle is: -12.22
sensed bank angle error is:   0.78 degrees
bank_angle_single_image_inferencing_elapsed_time: 0.084 seconds
```

Fig. 12. Artificial Interpolation vs Argmax Comparison

We can then gather up all the numerical angle fragments and sum them together.

```
bank_angle_array*prediction_vector:
   -0   -0   -0   -0   -0   -0   -0   -0   -0   -0
   -0   -0   -0   -0   -0   -0   -0   -0   -0   -0
   -0   -0   -0   -0 -0.00 -2.67 -9.85 -0.01 -0.00 -0
    0    0    0    0    0    0    0    0    0    0
    0    0    0    0    0 0.00    0    0    0    0
    0    0    0    0    0    0    0 0.00 0.01 0.00
    0
```

Fig. 13. Gathering Angle Fragments for Numerical Result

## V. HYPERPARAMETER OPTIMIZATION

*A. Introduction*

Early in developing the testbed, it appeared that a complex hyperparameter optimization using grid or random search would be necessary to obtain fast and accurate solutions. Paradoxically, so many iterations were required during testbed development that a good sense of the best values was brought while creating it. An interesting heuristic pattern arose in this process, which is outlined below.

*B. Iterative Thresholding*

There are two goals (at least) we can aspire to in computing the best value for hyperparameters.

Goal: Obtain the most accuracy in the shortest time.

Iterative thresholding is this: Optimize by selecting hyperparameter values that maximize the bank recognizer's train/test speed while maintaining adequate accuracy for practical use.

The term *num_train_copies* represents the number of occurrences of each symbol in the training set, enabling this optimization. It was found that accuracy was highly sensitive to the number of training set copies until a threshold was reached.

*C. Result: Five or Fewer Epochs are Needed for Best Train/Test Accuracy/Loss.*

For eleven runs of MNIST and Bank model train/test where accuracy was maximized for given *num_train_copies*, it was never necessary to exceed five epochs since improvement plateaued with more, and the extra computation was wasted. These models were always run together for verification reasons. A spurious result in the CNN for either MNIST or Bank flagged an error in the process and was very useful for development and debugging, in addition to increasing safety and reliability.

*D. Result: The Effect of Copy Number on Test Accuracy and Loss*

Below a specific *num_train_copies* threshold, testbed accuracy decreases due to the neural network's need for a minimum number of random rehearsals to recognize symbols. By employing binary search, the optimal *num_train_copies* is determined efficiently. For a typical run, the Bank Case requires only 20 copies (200 images) for quick, high-accuracy recognition, while MNIST necessitates 6,000 copies (60,000 images) due to handwriting variability. This is not surprising since the shape of the indicator glyph is fixed, except for rotation.

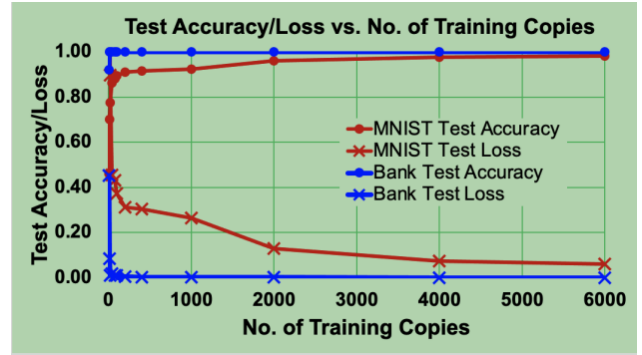

Fig. 14. Test Accuracy and Loss vs. Copy Number

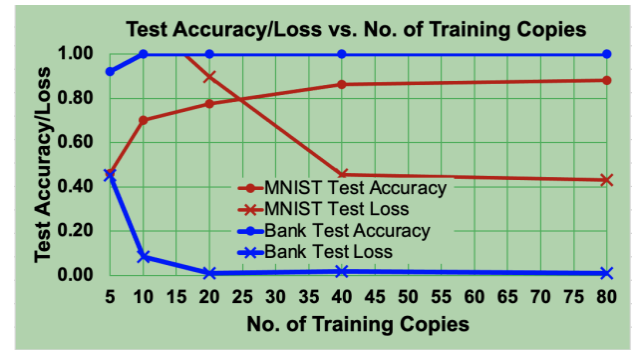

Fig. 15. Zoom into Accuracy Loss for Low Copy Number

*E. Result: The Effect of Copy Number on Train/Test Time*

We also see in the figures below the impact of *num_training_copies* on the training and testing time.

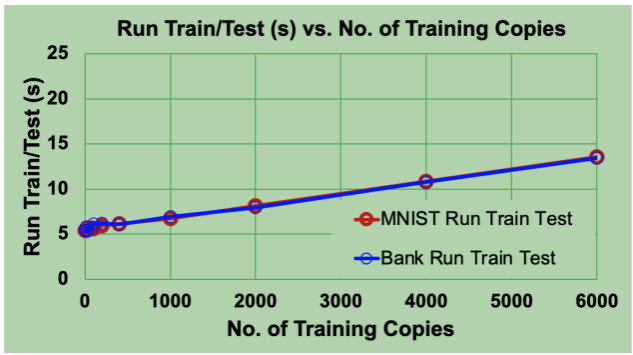

Fig. 16. Train/Test Time vs Copy Number

The first result is that the training and testing time for the model was the same for MNIST and Bank when they were processing the same number of images. Also, the training time is relatively short compared to the train/test image generation process, which will be discussed briefly.

*F. Running Down an Execution Time Anomaly*

Execution time anomalies sometimes occur, creating outliers in the measurement results. This happened during consecutive runs under differing run conditions, as shown in the figure below. The problem was not because the *num_training_copies* was increasing, but rather resource provisioning during the automated cloud GPU runs would vary depending on system load. The first run in a series, because of provisioning, RAM allocation, or GPU selection, differs significantly in elapsed time between the 1000 and 2000 copy number cases. This problem was examined in detail by doing five runs, discarding the first run, and averaging the remaining four. First runs averaged 22.5 ± 0.8 seconds and subsequent runs averaged 8.1 ± 0.3 seconds. Startup runs were typically the worst offenders, but this does not affect a deployed, pretrained system. Subsequent runs were used in the data above; below is just an example of a timing discrepancy.

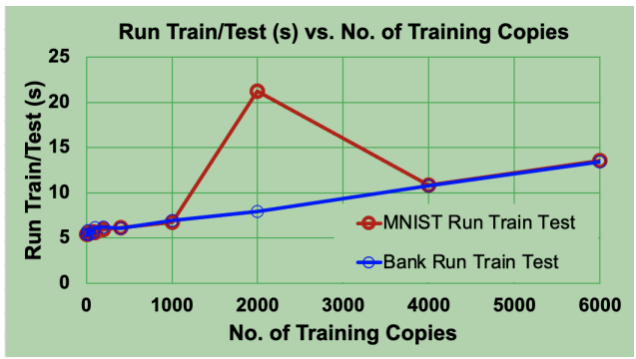

Fig. 17. Anomaly Between First and Subsequent Runs

*G. Result: Train/Test Time is Linear in Copy Number Selection*

Higher copy numbers increased training durations. Setting a minimum number of copies optimizes bank training performance, improving it by a factor of 2.8 in both MNIST and Bank scenarios.

*H. Result: The Effect of Copy Number on Model Inferencing Rate*

The graph below demonstrates the relationship between Model Inferencing Rate and num_train_copies, with the most significant changes for MNIST and Bank occurring within the initial 1,000 copies before plateauing. In the separate test recognizing 100 random images, rates exceed 1,500 FPS, with MNIST being around 25% slower than Bank.

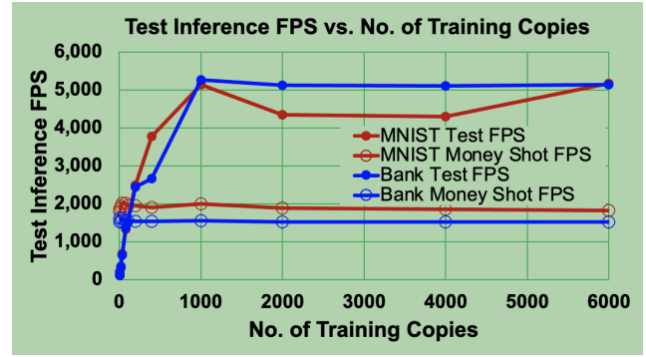

Fig. 18. Test Inferencing Rate vs. Copy Number

As the copy number graph exhibits substantial change early on, we focus on lower copy number region to identify the optimal copy number for achieving sufficient inferencing frame rates:

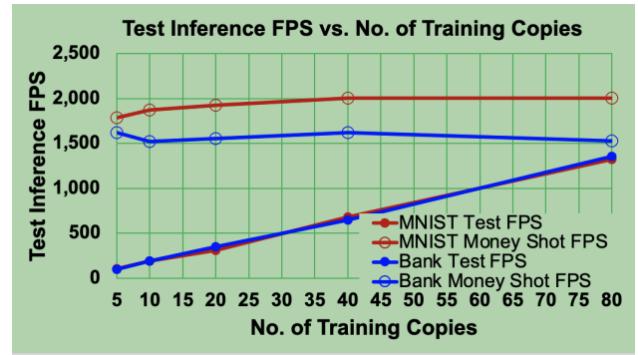

Fig. 19. Zoomed Inferencing Rate vs. Copy Number

A copy number of 20, previously found to be the minimum from accuracy considerations, offers a sufficient frame rate for bank inferencing.

*I. Result: Test/Train Image Generation Times vs Copy Number*

The MNIST train/test dataset is premade, while the Bank dataset requires generation from the canon. Because of this, MNIST generation time is unmeasurable. Since it is automated, bank image generation time is linearly proportional to the number of copies. Generating large train/test datasets is more time-consuming than training or inferencing times. However, generating large datasets for the Bank Instrument is unnecessary, as discovered in the accuracy results section.

*J. Generating Train/Test Images Once Saves Computer Time*

The following figure shows execution times for different sections of a Bank Instrument run. An import time saver MNIST

uses is splitting the train/test image generation from the inferencing task. Image generation currently takes more time than training, but it only needs to be done once in a deployed context.

| | TimeB |
|---|---|
| canon_gen_s | 2.514 |
| gen_train_test_s | 107.725 |
| run_train_test_s | 6.72 |
| model_save_s | 0.954 |
| test_infer_s | 0.982 |
| money_shot_infer_s | 0.06309 |
| single_image_infer_s | 0.212 |

Fig. 20. Times for a Typical Bank Run

## VI. RESULTS

The system achieves:

- 100% test accuracy on synthesized bank angles.
- Real-time frame rates exceeding 1000 fps.
- CNN interpolation for sub-degree angle resolution.
- 22X glyph image size reduction without accuracy loss.
- 5X training speedup through optimal hyperparameters.

## VII. INSIGHTS FOR ML SYSTEM DEVELOPMENT

Several insights emerged during system development:

- Generating train/test data is often the bottleneck vs model training.
- Careful task decomposition is key over long development cycles.
- Optimal label encoding took time to perfect.
- File path handling differed from OS vs. NumPy libraries.
- Fast edit/run/review cycles expedite hyperparameter optimization.
- Asking the right question can leapfrog months of effort.

The most impactful strategies were rational task decomposition and asking good questions.

## VIII. CONCLUSION

This work demonstrates a practical CNN approach for deducing aircraft state from images of cockpit instrumentation. It achieves real-time performance exceeding 1000 fps with high accuracy. The techniques introduced, including single image train/test data synthesis, CNN artificial interpolation, and the "Clean Training Principle," have broad applicability in ML system development. CNNs have incredible visual acuity.

The flexible approach enables interfacing robotic systems with existing instrumentation without costly upgrades. It provides a pathway for onboard vision-based aviation, powerplant, and factory-floor automation. This work combines proven ML techniques with an aircraft control metaphor to showcase valuable innovations in deep learning and computer vision.

ACKNOWLEDGMENT

The author gratefully acknowledges the advice and assistance of his dissertation committee, including Dr. Mariofanna Milanova, Dr. Dan Berleant, Dr. Ahmed Abu Halimeh, and Dr. Chia-Chu Chiang, in completing this work.